\documentclass[journal]{IEEEtran}
\usepackage{amsmath,amsfonts}
\usepackage[T1]{fontenc}
\usepackage{algorithmic}
\usepackage{algorithm}
\usepackage{array}
\usepackage[caption=false,font=normalsize,labelfont=sf,textfont=sf]{subfig}
\usepackage{textcomp}
\usepackage{stfloats}
\usepackage{url}
\usepackage{verbatim}
\usepackage{cite}
\usepackage[table]{xcolor}         
\usepackage[pdftex]{graphicx}
\usepackage{multirow}
\usepackage{multicol}
\usepackage{hyperref}
\begin{document}

\title{LEPA: Learning Geometric Equivariance in Satellite Remote Sensing Data with a Predictive Architecture}

\author{
  Erik~Scheurer,
  Rocco~Sedona, 
  Stefan~Kesselheim,
  Gabriele~Cavallaro
        
\thanks{Erik Scheurer is with the J\"{u}lich Supercomputing Centre (JSC), Forschungszentrum J\"{u}lich, 52428 J\"{u}lich, Germany and the Institute for Visualization and Interactive Systems (VIS),  University of Stuttgart, 70569 Stuttgart (e-mail: \href{mailto:e.scheurer@fz-juelich.de}{e.scheurer@fz-juelich.de)}}
\thanks{Rocco Sedona and Stefan Kesselheim are with the  J\"{u}lich Supercomputing Centre (JSC), Forschungszentrum J\"{u}lich, 52428 J\"{u}lich, Germany  (e-mail: r.sedona@fz-juelich.de; s.kesselheim@fz-juelich.de)}
\thanks{Gabriele Cavallaro is with the School of Engineering and Natural Sciences (SENS), University of Iceland, 102 Reykjavík, Iceland, and the J\"{u}lich Supercomputing Centre (JSC), Forschungszentrum J\"{u}lich, 52428 J\"{u}lich, Germany (e-mail: \href{mailto:gcavallaro@hi.is}{gcavallaro@hi.is}}
\thanks{This research is carried out as part of the Embed2Scale project and is co-funded by the EU Horizon Europe program under Grant Agreement No. 101131841. Additional funding for this project has been provided by the Swiss State Secretariat for Education, Research and Innovation (SERI) and UK Research and Innovation (UKRI). 
}
\thanks{The authors gratefully acknowledge the Gauss Centre for Supercomputing e.V. (www.gauss-centre.eu) for funding this project by providing computing time through the John von Neumann Institute for Computing (NIC) on the GCS Supercomputer JUWELS \cite{juwels} at Jülich Supercomputing Centre (JSC).}

}
\maketitle

\begin{abstract}

    Geospatial foundation models provide precomputed embeddings that serve as compact feature vectors for large-scale satellite remote sensing data. While these embeddings can reduce data-transfer bottlenecks and computational costs, Earth observation applications can still face geometric mismatches between user-defined areas of interest and the fixed precomputed embedding grid. 
    Standard latent-space interpolation is unreliable in this setting because the embedding manifold is highly non-convex, yielding representations that do not correspond to realistic inputs. 
    We verify this using Prithvi-EO-2.0 to understand the shortcomings of interpolation applied to patch embeddings. 
    As a substitute, we propose a Learned Equivariance-Predicting Architecture (LEPA). Instead of averaging vectors, LEPA conditions a predictor on geometric augmentations to predict the transformed embedding. We evaluate LEPA on NASA/USGS Harmonized Landsat-Sentinel (HLS) imagery and ImageNet-1k. Experiments show that standard interpolation has a mean reciprocal rank (MRR) below 0.2, whereas LEPA increases MRR to over 0.8, enabling accurate geometric adjustment without re-encoding.\end{abstract}


\section{Introduction}

\IEEEPARstart{T}{o} increase accessibility, speed up prototyping, and address computational and memory bottlenecks, embedding datasets are becoming common in Earth observation (EO) \cite{alphaearth, majortomembeddings, tesseraembeddings}. Instead of downloading and preprocessing terabytes of remote sensing data, embeddings, vectorial proxies of the data, are precomputed with foundation models and yield more expressive and compact representations compared to the raw inputs \cite{prithvi2, terramind, alphaearth}. However, users may face geometric mismatches between user-defined areas of interest and the fixed precomputed embedding grid. It is desirable to directly transform embeddings to match user data to avoid repeated expensive encoder inference passes after an alignment in image space. While image interpolation is well-studied \cite{interpolation}, geometric operations on embeddings, especially patch embeddings, are unreliable: latent-space structure depends on the foundation model and can be highly non-convex \cite{latentcartography}. 

Recently, joint-embedding predictive architectures (JEPAs) gained traction in computer vision \cite{ijepa, lejepa} and remote sensing \cite{rejepa,anysat, galileo, sarjepa}.
These methods append a predictive model to an encoder during training to undo a transformation such as masking or color changes in embeddings directly \cite{ijepa, iwm}. This information bottleneck forces the encoder to learn meaningful properties in the inputs.
Usually, the predictor is discarded after pretraining. Instead, we propose to use the predictor to predict the geometric transformations on embeddings to approximate geometric equivariance.
If a transformation $t$ in embedding space of encoder $E$ satisfies $t(E(x)) = E(T(x))$ for an image-space transformation $T$, then $E$ is \emph{equivariant} to $T$.
Approximating such equivariance is an active research area \cite{groupequivariantcnn, steerabletransformersforvolume, eqvae, care, steerableequivlearning, iwm}, though most approaches have not been scaled, are not steerable, or use non-geometric transformations. To address these limitations, we train an I-JEPA \cite{ijepa} model on ImageNet-1k \cite{imagenet1k} and HLS data \cite{prithvi2} as a baseline and evaluate it on PANGAEA \cite{pangaea}. We show that an unmodified architecture already yields competitive embeddings. We then improve equivariance through a pretraining strategy that enforces predictor sensitivity to geometric transformations, inspired by image world models (IWM) \cite{iwm}. 

To test equivariance, the mean reciprocal rank (MRR) is evaluated on Prithvi-EO-2.0, I-JEPA and our JEPA variant together with the world model, which we call a learned equivariance-predicting architecture (LEPA). We find that standard interpolation on patch-embeddings is surpassed substantially by the learned approach with further gains from equivariance fine-tuning.
Finally, we analyze the embeddings to compare datasets and adjust the model architecture to further enhance equivariance in I-JEPA models.
Our contributions can be summarized as follows.
\begin{enumerate}
    \item We find that traditional interpolation and downsampling methods break down when used with patch embeddings.
    \item For efficiently aligning embeddings to user data, we propose to avoid repeated inference passes of large foundation models with an additional predictive model.
    \item We introduce an I-JEPA model trained on either ImageNet or HLS data that performs competitively on PANGAEA \cite{pangaea} along with architectural improvements such as a classification token and novel positional encodings.
\end{enumerate}

\begin{figure}
    \centering
    \includegraphics[width=\linewidth]{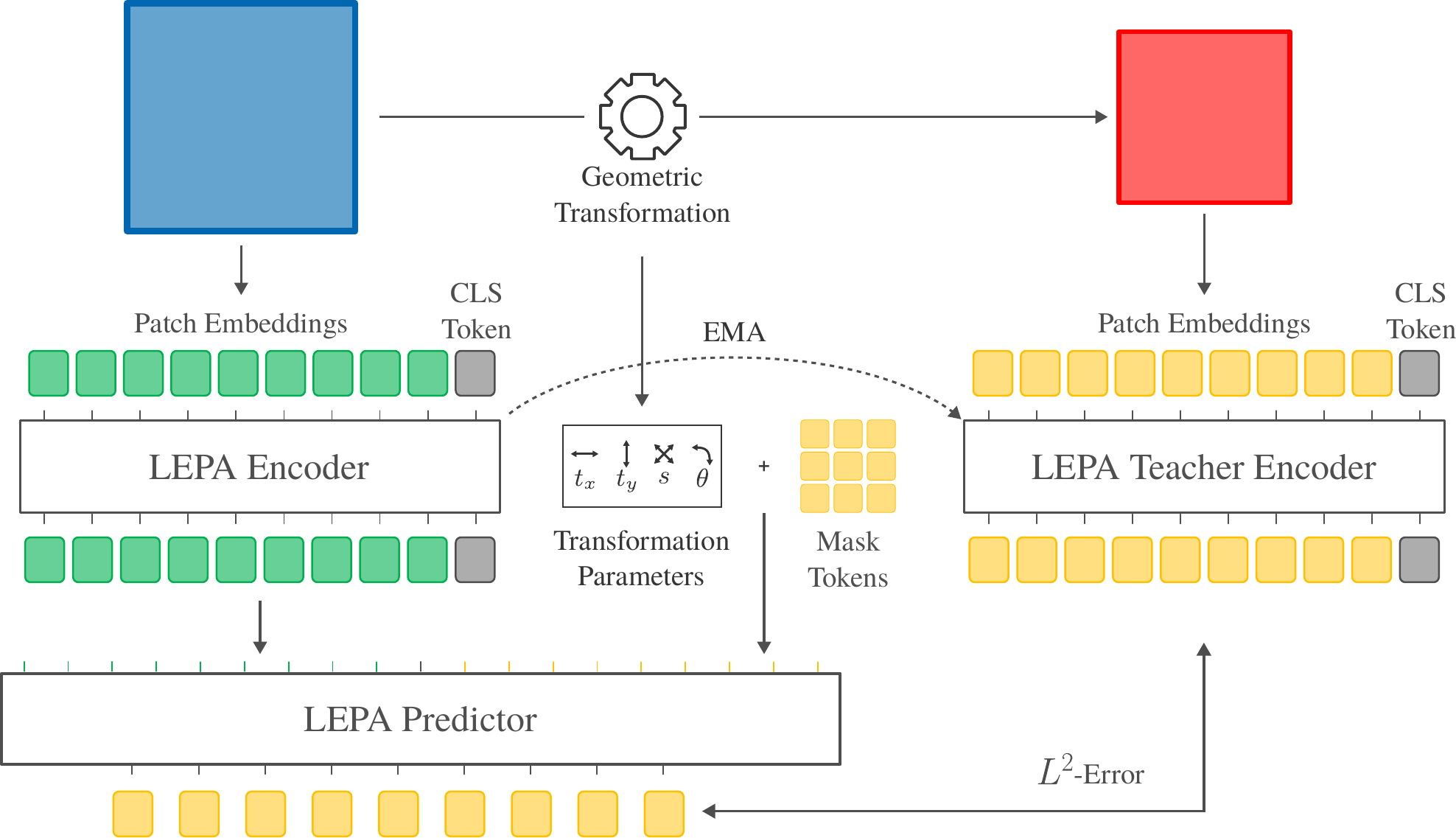}
    \caption{LEPA training architecture: the original input is passed to the student encoder which produces patch embeddings. These patch embeddings form the context to the predictor along with transformation parameters. Predicted embeddings are compared against the teacher output which are computed from a transformed input image. Masking is omitted for visual clarity.}
    \label{fig:architecture}
\end{figure}
\section{Method}
\paragraph{Equivariance of Prithvi-EO-2.0}
We test the equivariance of Prithvi-EO-2.0-300M \cite{prithvi2} as sample MAE architecture by transforming the produced patch embeddings. The patch embeddings form a grid which is either rearranged according to a 90-degree counterclockwise rotation or downsampled to half resolution using bilinear interpolation. The patch embeddings are reconstructed using a fine-tuned decoder that reconstructs non-masked patch embeddings into image space.\paragraph{I-JEPA} The pretraining task of the image-based joint-embedding predictive architecture (I-JEPA) \cite{ijepa} is latent inpainting. A teacher produces representations from the full image, and blocks of patches are selected as targets. A ViT encoder processes a non-overlapping region to provide context to the key novelty of I-JEPA: a predictor that inpaints missing blocks conditioned on the context via cross-attention. The teacher is an exponentially moving average of the student. A follow-up, image world models \cite{iwm}, introduces an additional pretraining task to enforce equivariance: given an augmented context and augmentation parameters, predict target embeddings generated from the unmodified image. The predictor thus learns both latent inpainting and approximate reversal of augmentations such as color jitter, grayscale, and contrast changes, producing embeddings equivariant to these augmentations \cite{iwm}. We adapt this framework to geometric transformations.
\paragraph{LEPA} Geometric transformations such as rescaling, rotation and translation are not trivially defined for patch-wise embeddings, since parts of a patch move to neighboring patches under these operations. A linear combination of embeddings, as in interpolation, may not yield a valid vector in a highly non-convex embedding manifold. Thus the predictor must capture both spatial information within each patch-embedding and the desired geometric transformation. The full training architecture is displayed in Fig.~\ref{fig:architecture}. Training pairs are obtained by transforming the unmodified context (blue) into a target image (red) using translation in $x$/$y$, rotation, and scaling. These parameters are appended to the predictor’s mask-token and passed through a 3-layer MLP projecting back to the embedding dimension, enabling the predictor to inpaint missing patches while performing the transformation in embedding space (yellow) given the encoded context (green). For better performance, the full target image is predicted rather than blocks as in I-JEPA. We also test an inductive bias via novel conditioned positional encodings. The height and width positional indices of each patch are centered around the image center rather than starting at a corner. A formal definition and illustration are provided in the appendix. 
Through this switch, the positional encodings can reflect the changed position of the patch positions under transformation.
\section{Experiments}

\subsection{Transforming Prithvi-EO-2.0 embeddings}
\begin{figure} 
\footnotesize
    \centering
    Rotation\\
    \includegraphics[width=0.8\linewidth]{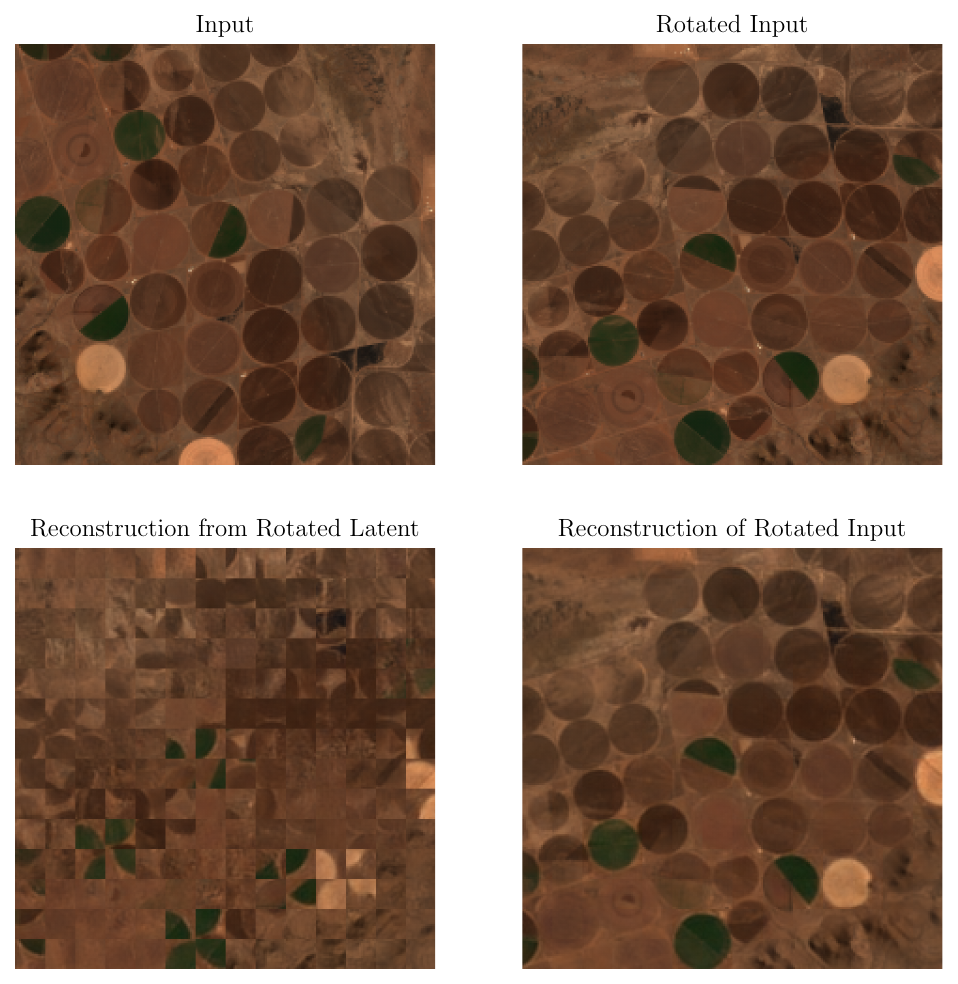}\\
    Downsampling\\
    \includegraphics[width=0.8\linewidth]{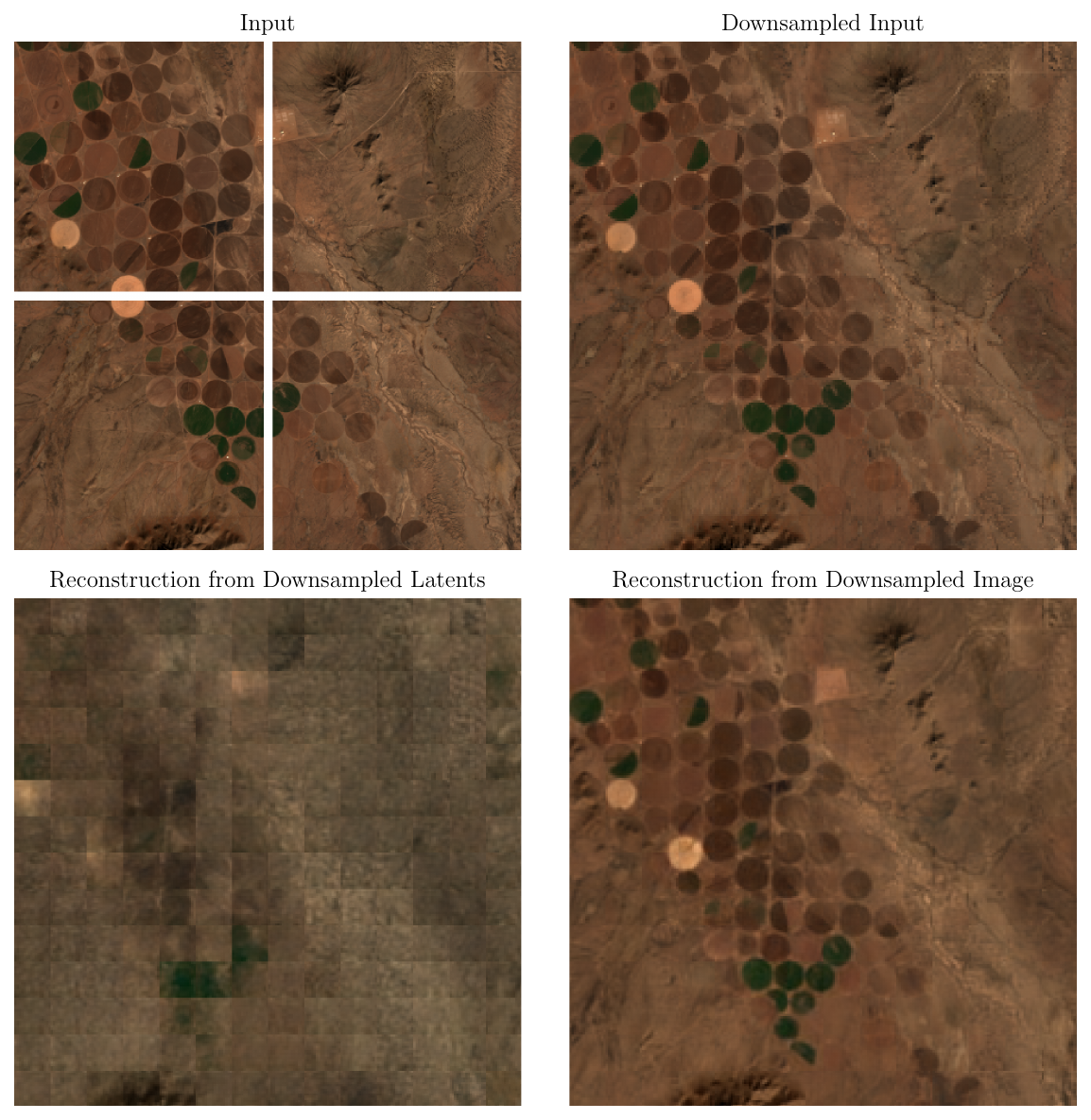}
    \caption{Input and reconstructions of rotated and downsampled images using a finetuned version of Prithvi \cite{prithvi2}. The bottom left images depict reconstructions from a transformed latent, the bottom right images are reconstructions from rotations in image space as a baseline.}
    \label{fig:prithvi-recons}
\end{figure}

We test rotating in embedding space instead of image space and reconstructing both the rotated latents and the rotated image. The reconstructions are shown in Fig.~\ref{fig:prithvi-recons}. 
For rotated latents, each patch is clearly visible in the reconstruction. While the patches remain the same as in the input, the relation of the different patches changes due to the rotation. Because MAE patch embeddings contain information on the geometric structures within the patch to allow for reconstruction and inpainting, changing the arrangement of embedding vectors does not change the content of these vectors. 
Downsampling representations using bilinear interpolation reveals the non-convex nature of the embeddings. Since a patch embedding saves pixel arrangements, na\"ively combining different vectors does not yield meaningful embeddings. This experiment calls into question whether embedding vectors of larger areas can be averaged for further inference, e.g.\ classification of an area of multiple patches. Without specific testing, it must not be assumed that embedding vectors can be linearly combined.

\subsection{Model Training}
Both I-JEPA and LEPA train for 50 epochs either on ImageNet-1k \cite{imagenet1k} or the HLS dataset used for pretraining Prithvi-EO-2.0 \cite{prithvi2} based on a ViT-base architecture \cite{vit}. We publish the code under \url{https://github.com/embed2scale/LEPA}. A performance analysis reveals that 50 epochs are sufficient for performance convergence even though the representations continue to change. The number of epochs is chosen to decrease training time and to avoid overfitting to the data. With longer training the noise discussed in Sec.~\ref{sec:equivariance} increases and therefore the equivariance decreases.

\subsection{Representation Quality}

The PANGAEA \cite{pangaea} benchmark is used to measure embedding quality. For each dataset in PANGAEA a UPerNet decoder \cite{upernet} is trained to map from frozen encoder representation to semantic segmentation map. Despite UPerNets's design, no intermediate outputs are passed to the decoder to imitate a fixed embedding dataset. As a baseline, we compare Prithvi-EO-2.0 \cite{prithvi2}, TerraMind \cite{terramind}, RemoteCLIP \cite{remoteclip}, DOFA \cite{dofa} and CROMA \cite{croma} with our JEPA variants. For a more global representation and visual clarity we compute a normalized score for each model by normalizing the score for each dataset using the baselines and two I-JEPA models -- one of each pretraining dataset. The performance statistics are computed for each model and displayed in Fig.~\ref{fig:normalized-score}. The full benchmark table is in the supplementary material.
We find that JEPA models perform competitively without architectural changes. While TerraMind-B outperforms all variations in most datasets, their dataset is much larger and can use multimodal information in cases where Sentinel-1 information exists, i.e.\ Sen1Floods11, CropTypeMapping, and PASTIS.

\paragraph{Impact of Dataset}
When training with a JEPA target, we obtain high-quality embeddings. Pretraining data strongly influences downstream performance: ImageNet-1k-pretrained I-JEPA achieves competitive scores despite being out-of-distribution. Unlike ImageNet’s fixed, class-balanced sampling, HLS is stratified across land-cover classes and ecoregions, with entropy-based tiles for heterogeneous areas. Temporal sequences (four seasonal timestamps) were extracted per tile, filtered for clouds, and split into patches for spatial and temporal diversity \cite{prithvi2}. Thus, ImageNet emphasizes single-object classification, while HLS captures diverse landscapes. The ImageNet variant of I-JEPA outperforms others on the Marine Debris and Oil Spill (MADOS) dataset, which is challenging due to 15 classes and skewed distribution \cite{mados}. The focus of ImageNet on classes likely aids detection. ImageNet variants also exceed HLS when no multispectral data are available.

\paragraph{Architecture Modifications} Prepending a classification (CLS) token  to the patch embeddings to aggregate global information \cite{vit} improves semantic segmentation for ImageNet models across datasets (except AI4SmallFarms), even though only patch embeddings feed the decoder. HLS models show no consistent effect when adding a CLS-token, suggesting differences lie within decoder variance.

Modified positional encodings and the auxiliary transformation prediction objective had no measurable impact, confirming earlier findings \cite{vit} that positional encoding choice matters less for embedding quality than their presence.

\begin{figure}
    \centering
    \includegraphics[width=\linewidth]{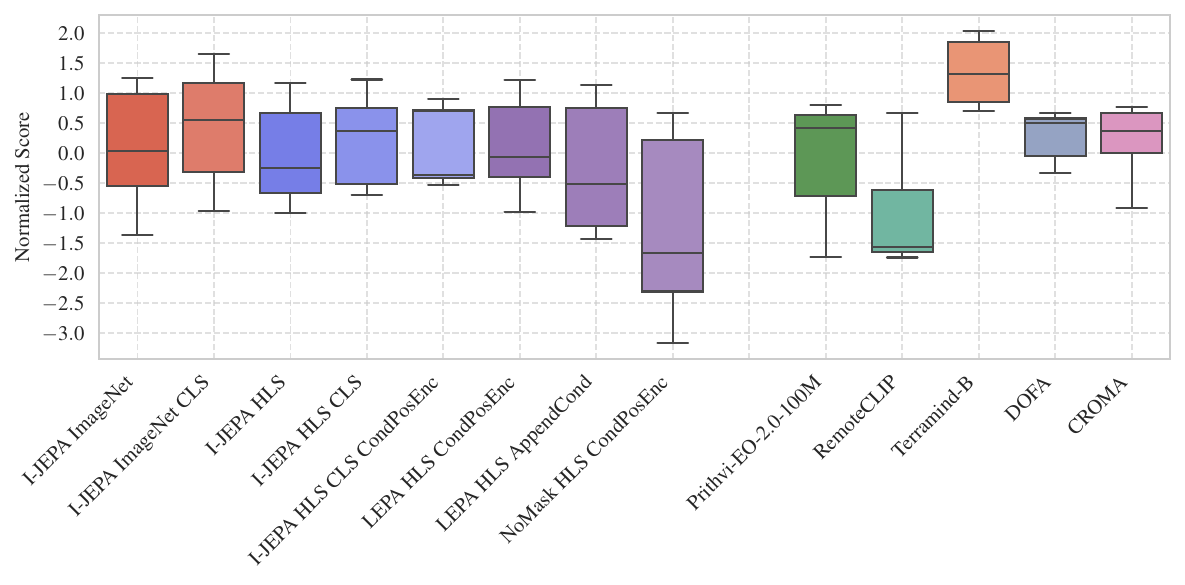}
    \caption{Box-plot of the normalized score based on the results of Table~\ref{tab:mrr}.}
    \label{fig:normalized-score}
\end{figure}

\subsection{Representation Equivariance}
\label{sec:equivariance}
We evaluate the equivariance of representations using the Mean Reciprocal Rank (MRR) \cite{iwm, mrr}. For each image, a series of 256 different augmentations is computed. Then an unmodified version of the image is passed through the encoder and predictor trying to predict the embeddings of one of the previous transformations. The different augmentations are sorted by cosine similarity to the predicted embedding.

The mean reciprocal rank is computed by averaging these ranks as
\begin{equation}
    \text{MRR} = \frac{1}{N}\sum^N_{n=1} \frac{1}{\text{rank}_n},
\end{equation}
where $\text{rank}_n$ is the rank of the $n$-th sample and N the number of images in the evaluation set. This metric is a useful proxy to the capacities of the model, as it can differentiate between invariance and equivariance. 
An invariant model produces similar embeddings for each augmentation, yielding an unstable rank and thus a lower MRR. This makes the metric more informative than a simple $L^2$ distance or cosine similarity between target and predicted embeddings, which are also small for invariant models. For encoders without a predictor, the augmented embedding is computed by interpolating the patch embeddings using nearest neighbors. Interpolation using bilinear interpolation yielded slightly lower MRR. This finding confirms the loss of information when linearly combining embeddings. As baseline, Prithvi-EO-2.0 is used as it is trained on the same HLS dataset as our I-JEPA models. Adding more models as a baseline requires a different dataset or padding the model inputs, both of which might introduce new biases into the evaluation.

\begin{figure}[t]
    \centering
    \setlength{\tabcolsep}{1pt} 
    
    \begin{tabular}{c ccc cc cccccc}
        
                                                                                  & \multicolumn{3}{c}{ImageNet} & \multicolumn{2}{c}{HLS} \\[0pt] 
        \raisebox{.87\height}{\rotatebox[]{90}{\footnotesize Input Image}}          & \hspace{-3pt}
        \includegraphics[width=.159\linewidth]{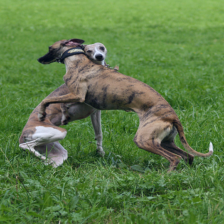}        &
        
        \includegraphics[width=.159\linewidth]{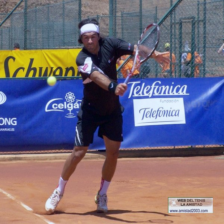}        &
        
        \includegraphics[width=.159\linewidth]{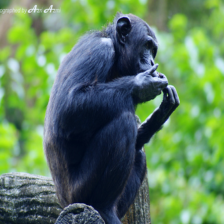}        & \hspace{5pt}
        \includegraphics[width=.159\linewidth]{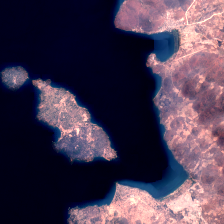}          &
        
        \includegraphics[width=.159\linewidth]{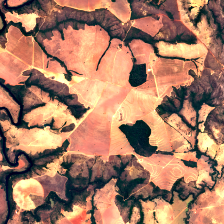}          &

        \\ 

        \raisebox{1\height}{\rotatebox[]{90}{\footnotesize w/o CLS}}                & \hspace{-3pt}
        \includegraphics[width=.159\linewidth]{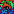} &
        
        \includegraphics[width=.159\linewidth]{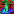} &
        
        \includegraphics[width=.159\linewidth]{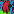} & \hspace{5pt}
        \includegraphics[width=.159\linewidth]{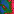}   &
        
        \includegraphics[width=.159\linewidth]{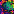}   &

        \\ 

        \raisebox{1.2\height}{\rotatebox[]{90}{\footnotesize w/ CLS}}                 & \hspace{-3pt}
        \includegraphics[width=.159\linewidth]{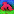}    &
        
        \includegraphics[width=.159\linewidth]{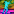}    &
        
        \includegraphics[width=.159\linewidth]{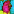}    & \hspace{5pt}
        \includegraphics[width=.159\linewidth]{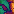}      &
        
        \includegraphics[width=.159\linewidth]{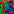}      &

        \\ 
        
    \end{tabular}
    \caption{Example images with class-specific noise. Bottom rows show the first two PCA components (per image) mapped to a color wheel. In the ImageNet model without a CLS-token, background embeddings resemble the subject, a pattern not seen consistently in HLS data.}
    \label{fig:class-noise}
\end{figure}

\paragraph{Embedding analysis} When analyzing the embeddings of the ImageNet model, a class-specific noise is noticeable, as seen in the background of Fig.~\ref{fig:class-noise}. This noise resembles that observed in DINOv3 \cite{dinov3}, since regions with different content show high similarity to parts containing the main subject, and because the effect increases with longer training. Adding a CLS-token reduces or removes these artifacts.
The HLS model, not trained on a fixed set of concepts and lacking a central subject, does not exhibit the same artifact pattern. However, artifacts still appear (Fig.~\ref{fig:class-noise}) and do not fully disappear even with a CLS-token, which is reflected in the MRR score in Tab.~\ref{tab:mrr}. A CLS-token improves interpolation equivariance for ImageNet models but not for HLS I-JEPA models, and for LEPA it even decreases MRR. We hypothesize that the open-ended nature of HLS data prevents useful global image information from being concentrated in a single CLS-token.

\begin{figure}[t]
    \centering
    \setlength{\tabcolsep}{1pt} 
    
    \begin{tabular}{c ccc cc}
        
        &-30°&-15°&0°&15°&30°\\
        \raisebox{.9\height}{\rotatebox[]{90}{\footnotesize Input Image}}          & \hspace{-3pt}
        \includegraphics[width=.159\linewidth]{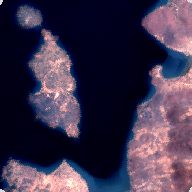}        &
        \includegraphics[width=.159\linewidth]{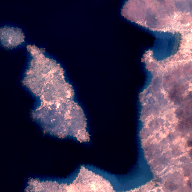}        &
        \includegraphics[width=.159\linewidth]{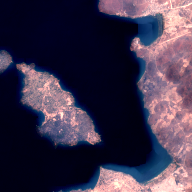}        &
        \includegraphics[width=.159\linewidth]{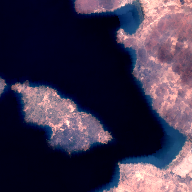}        &
        \includegraphics[width=.159\linewidth]{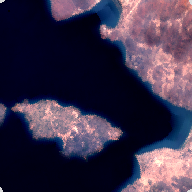}        
        \\ 
        \raisebox{1.5\height}{\rotatebox[]{90}{\footnotesize Targets}}          & \hspace{-3pt}
        \includegraphics[width=.159\linewidth]{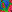}        &
        \includegraphics[width=.159\linewidth]{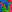}        &
        \includegraphics[width=.159\linewidth]{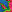}        &
        \includegraphics[width=.159\linewidth]{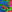}        &
        \includegraphics[width=.159\linewidth]{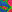}        
        \\ 

        \raisebox{.91\height}{\rotatebox[]{90}{\footnotesize Predictions}}          & \hspace{-3pt}
        \includegraphics[width=.159\linewidth]{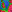}        &
        \includegraphics[width=.159\linewidth]{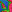}        &
        \includegraphics[width=.159\linewidth]{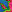}        &
        \includegraphics[width=.159\linewidth]{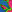}        &
        \includegraphics[width=.159\linewidth]{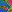}        
        \\ 

        \raisebox{.84\height}{\rotatebox[]{90}{\footnotesize Interpolation}}          & \hspace{-3pt}
        \includegraphics[width=.159\linewidth]{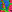}        &
        \includegraphics[width=.159\linewidth]{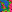}        &
        \includegraphics[width=.159\linewidth]{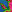}        &
        \includegraphics[width=.159\linewidth]{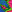}        &
        \includegraphics[width=.159\linewidth]{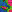}        
        \\ 
        
    \end{tabular}
    \caption{Angle sweep of a selected image comparing the first two PCA components using interpolation in image space (Targets), the predictions using finetuned LEPA, and nearest neighbor interpolation.}
    \label{fig:angle_sweep}
\end{figure}

\paragraph{Predicting augmentations} Conditioning the predictor on augmentation parameters markedly improves equivariance. With LEPA, the MRR increases from around 0.2 to nearly 0.7. Further finetuning of the predictor, while keeping the encoder frozen and training it solely to predict augmentations without inpainting, pushes the score to 0.8. We see the qualitative difference in transformation in Fig.~\ref{fig:angle_sweep} where we compare the augmentations done in image space, predicted using a LEPA predictor, and nearest neighbor interpolation using the same PCA decomposition.
This setup illustrates how the predictor can be deployed to adjust a grid of embeddings for sampling specific points. Image interpolation techniques have notable artifacts whereas the predicted embeddings stay close to the target embeddings.For training, we cannot use augmentation prediction as the exclusive objective, as shown by the no-masking training in Tab.~\ref{tab:mrr}: it underperforms in semantic segmentation (Tab.~\ref{tab:pangaea_results}) and does not necessarily yield a high MRR score. With this single objective, training becomes unstable, and the low equivariance is due to collapse into simpler invariant embeddings. Conditioning through positional encodings produces more equivariant embeddings than merely appending the parameters to the mask tokens and passing them through an MLP; after finetuning, however, the positional encodings reduce embedding equivariance. 

This switch raises questions on the conditioning on transformation parameters. When predicting transformation and inpainting at once, the inductive bias of conditional positional encodings improves predictions. When solely focused on prediction, increasing model complexity to understand the numerical values yields better results. This implies a suboptimal conditioning through the positional encodings. More consistent inductive biases might improve prediction performance further.

\begin{table}[t]
    \centering

    \caption{MRR of models given different hyperparameters with and without finetuning of the predictor. For Prithvi and I-JEPA, the predicted embedding is obtained with nearest-neighbor interpolation. Other interpolation methods decrease MRR.} 
    \begin{tabular}{l l c l c c }
        \hline
        \textbf{Model}                            & \textbf{Dataset} & \textbf{CLS} & \textbf{Pos. Enc.} & \textbf{MRR}    & \textbf{After Finetune} \\ \hline
        I-JEPA                                    & IN-1k            & No           & default         & 0.1837          & N/A                    \\
        \rowcolor{gray!20}I-JEPA                  & IN-1k            & Yes          & default         & 0.2108          & N/A                    \\
        I-JEPA                                    & HLS              & No           & default         & 0.1826          & N/A                    \\
        \rowcolor{gray!20}I-JEPA                  & HLS              & Yes          & default         & 0.1839          & N/A                    \\
        LEPA                                   & HLS              & No           & CondPos         & \textbf{0.6975} & 0.8062                 \\
        \rowcolor{gray!20}LEPA                 & HLS              & Yes          & CondPos         & 0.6630          & 0.7994                 \\
        LEPA                                   & HLS              & No           & default         & 0.6183          & \textbf{0.8355}        \\
        \rowcolor{gray!20}NoMask                  & HLS              & No           & CondPos         & 0.5906          & N/A                    \\
        \hline
        \multicolumn{4}{ l }{Prithvi-EO-2.0-100M} & 0.2113           & N/A                                                                       \\ \hline
    \end{tabular}
    \label{tab:mrr}
\end{table}

\section{Conclusion}
In this work we provide perspective on an inherent limitation of patch embeddings and find an alternative to image interpolation to transform patch embeddings using a learned world model. As a single vector encodes information on a patch of pixels, the geometric information is contained within and must be considered when using embeddings.
We provide way to avoid repeated encoder passes with our LEPA model. By extending the I-JEPA training target, we achieve embedding performance in line with other foundation models in addition to a predictor that understands the embedding space of the encoder. This predictor can be conditioned to predict image space transformations on patch embeddings directly to avoid repeated encoder passes with precomputed embeddings. Using this learned transformation, the MRR increases from $<0.2$ to $>0.8$, marking a substantial improvement transformation accuracy over nearest neighbor and bilinear interpolation.

We identify several directions to strengthen evaluation, equivariance and performance. First, additional foundation models can be probed for equivariance. Since prior work \cite{3diebench, steerableequivlearning, laligan} shows that linear predictors often suffice, a smaller predictor for LEPA and other models may reduce inference cost. Second, LEPA’s inductive bias could be improved with better conditioning such as relative positional encodings, e.g.\ ALiBi \cite{alibi} or RoPE \cite{rope}. For improving downstream performance, many opportunities are possible through regularization and changing the prediction task \cite{rejepa,lejepa,alphaearth}. Third, we will test whether a fixed number of classes induces the observed noise by training one model on a classification-oriented EO dataset and another on a general-purpose RGB dataset.

\bibliography{bibliography}

\newpage
\appendix
\paragraph{Positional Encodings}
\label{app:pos_encodings}

\begin{figure}[h]
    \centering
    \includegraphics[width=\linewidth]{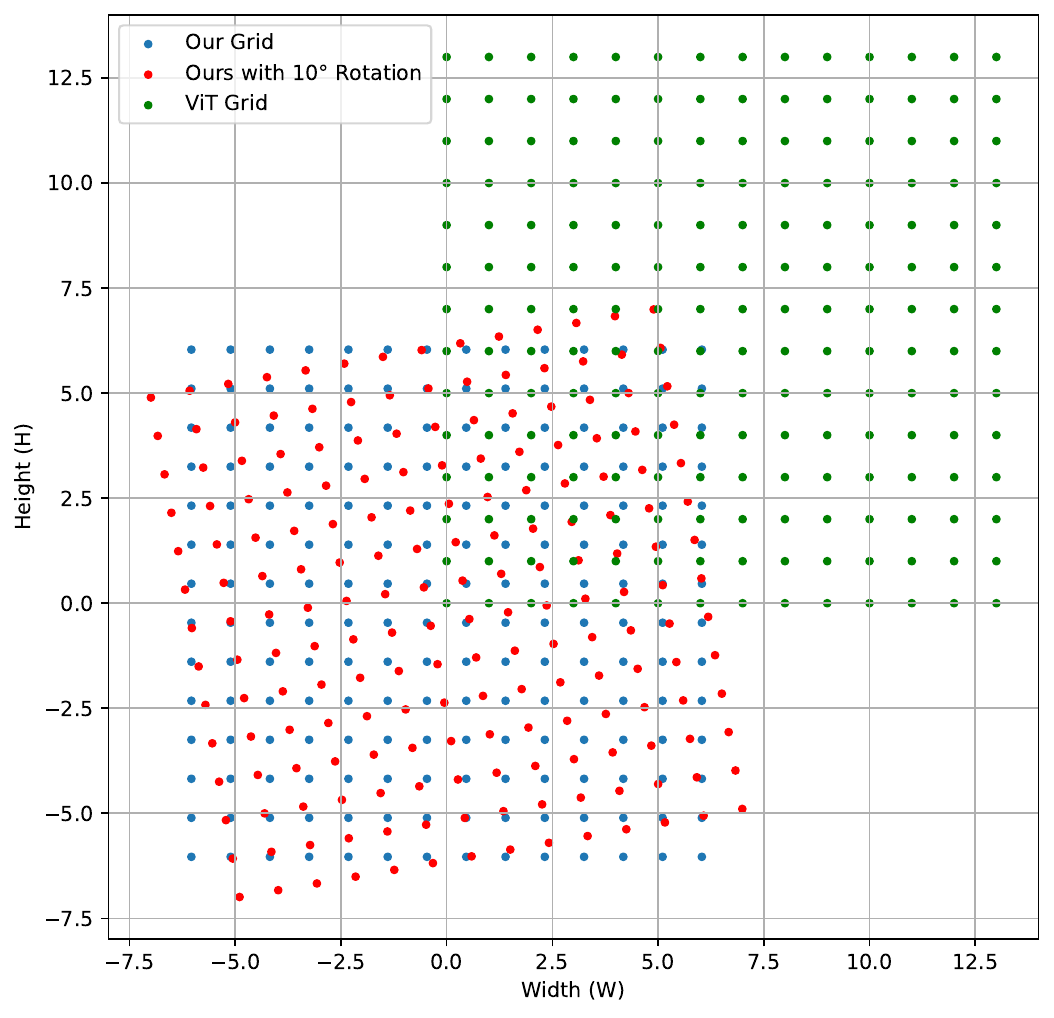}
    \caption{Grid that is used for sinusoidal positional encodings by default, ours without a transformation and ours with a rotation angle of 10°.}
    \label{fig:posenc_grid}
\end{figure}

\begin{figure}
    \small
    \centering
    Default
    \includegraphics[width=\linewidth]{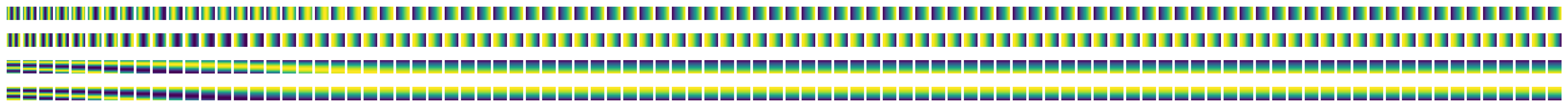}
    Our unmodified encodings
    \includegraphics[width=\linewidth]{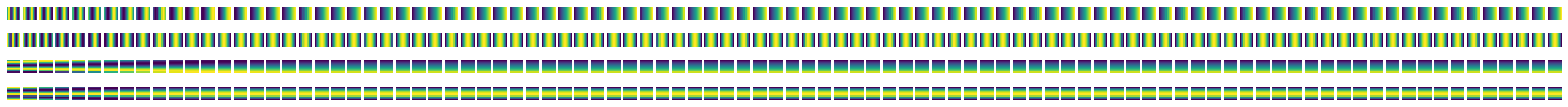}
    Our encodings rotated by 45°
    \includegraphics[width=\linewidth]{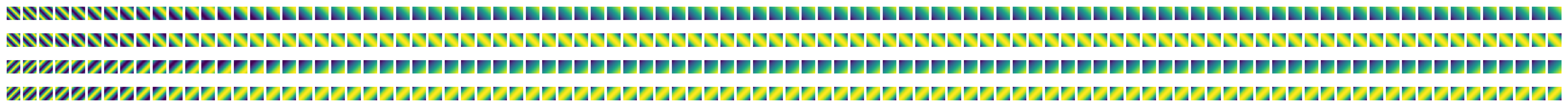}
    \caption{Comparison of positional encodings for each embedding dimension out of 768. Each row consists of one line of equation~\ref{eq:posenc}.}
    \label{fig:posenc}
\end{figure}

2D positional encodings of a standard ViT \cite{vit} are sinusoidal encodings for height and width position of each patch. The full embedding dimension is split into two parts as follows:
\begin{equation}
    \begin{aligned}
        PE_{(pos_h,2i)}   & = \sin(pos_h/10000^{2i/d_{model}} ), \\
        PE_{(pos_h,2i+1)} & = \cos(pos_h/10000^{2i/d_{model}} ), \\
        PE_{(pos_w,2i)}   & = \sin(pos_w/10000^{2i/d_{model}} ), \\
        PE_{(pos_w,2i+1)} & = \cos(pos_w/10000^{2i/d_{model}} ),
    \end{aligned}
    \label{eq:posenc}
\end{equation}
with $pos_w$ and $pos_h$ denoting the $x$- and $y$-coordinates in a grid and $i=0...D/2$ for embedding dimension $D$. These positional encodings are then concatenated into a single embedding vector. For our conditioned embeddings we only change the underlying grid that is used to obtain the positions. We show the different grids used to index the patches in Fig.~\ref{fig:posenc_grid}. Our grid is centered at the origin of (0,0) instead of the edge of an image. With this modification, changing the angle, translation and scaling can easily be realized. These conditioned positional encodings are added to the mask-tokens in the same way as other absolute positional encodings. The generated positional encodings for each embedding dimension are shown in Fig.~\ref{fig:posenc}.

\newpage
\paragraph{PANGAEA benchmark table}

\begin{table*}[t]
    \centering
    \caption{Mean IoU comparison of pretrained foundation models and JEPA variations of the PANGAEA benchmark \cite{pangaea}. The \textbf{best} and \underline{second best} models are marked for each dataset.}
    \label{tab:pangaea_results}
    \resizebox{\linewidth}{!}{
        \begin{tabular}{ l l c l c c c c c c c c c }
            \hline
            \textbf{Model}                                                              & \textbf{Dataset} & \textbf{CLS}   & \textbf{PosEnc} & \textbf{Params}   & \textbf{HLSBurnscars} & \textbf{Sen1Floods11} & \textbf{MADOS}    & \textbf{AI4SmallFarms} & \textbf{PASTIS}   & \textbf{DynamicEarthNet} & \textbf{SpaceNet7} \\
            \hline
            I-JEPA                                                                      & IN-1k            & No             & default         & 85.8M             & 77.15                 & 71.75                 & 59.28             & \underline{26.58}      & 21.31             & 35.13                    & 60.02              \\
            \rowcolor{gray!20} I-JEPA                                                   & IN-1k            & Yes            & default         & 85.8M             & 77.58                 & 74.74                 & \underline{65.59}    & 25.98                  & 24.07             & 37.68                    & \textbf{61.12}  \\
            I-JEPA                                                                      & HLS              & No             & default         & 86.4M             & 82.82                 & 86.69                 & 45.74             & 24.33                  & 33.00             & 29.95                    & 56.93              \\
            \rowcolor{gray!20} I-JEPA                                                   & HLS              & Yes            & default         & 86.4M             & \textbf{83.08}     & 85.87                 & 46.40             & 25.48                  & \underline{35.03} & 31.46                    & 56.67              \\
            I-JEPA                                                                      & HLS              & Yes            & CondPos         & 86.4M             & 81.73                 & 85.88                 & 47.59             & 24.58                  & 34.43             & 32.64                    & 56.64              \\
            \rowcolor{gray!20} LEPA                                                  & HLS              & No             & CondPos         & 86.4M             & \underline{83.03}                 & 87.29                 & 51.40             & 23.82                  & 33.84             & 33.84                    & 56.17              \\
            LEPA                                                                     & HLS              & Yes            & CondPos         & 86.4M             & 82.71                 & 85.44                 & 45.14             & 24.01                  & 33.77             & 32.13                    & 55.61              \\
            \rowcolor{gray!20} LEPA                                                  & HLS              & No             & default         & 86.4M             & 82.72                 & \underline{87.37}     & 43.36             & 23.26                  & 33.61             & 28.06                    & 56.31              \\
            NoMask                                                                      & HLS              & No             & CondPos         & 86.4M             & 80.81                 & 83.89                 & 33.51             & 21.12                  & 29.12             & 26.65                    & 51.96              \\
            \hline
            \rowcolor{gray!20}\multicolumn{4}{ l }{Prithvi-EO-2.0-100M \cite{prithvi2}} & 86.4M            & 80.28          & 87.12           & 41.46             & 26.03                 & 31.12                 & 26.27             & 57.12                                                                                      \\
            \multicolumn{4}{ l }{RemoteCLIP \cite{remoteclip}}                          & 87.5M            & 70.98          & 70.28           & 53.11             & 23.09                 & 15.01                 & \underline{38.28} & 54.33                                                                                      \\
            
            \rowcolor{gray!20} \multicolumn{4}{ l }{TerraMind-B \cite{terramind}} & 87.7M & 81.40 & \textbf{88.42 }& \textbf{67.44} & \textbf{27.55} & \textbf{41.53} & \textbf{38.46} & \underline{60.56} \\
            \multicolumn{4}{ l }{DOFA \cite{dofa}}                                      & 111.3M           & 76.63          & 85.51           & 50.32             & 25.70                 & 28.81                 & 38.00             & 59.04                                                                                      \\
            \rowcolor{gray!20}\multicolumn{4}{ l }{CROMA \cite{croma} }                 & 201.5M           & 79.56          & 87.26           & 54.97             & 23.90                 & 33.72                 & 37.97             & 56.66                                                                                      \\

            \hline
        \end{tabular}
    }
\end{table*}

Table~\ref{tab:pangaea_results} displays the score for each dataset in PANGAEA \cite{pangaea} individually along with the parameter count of the corresponding model. Our JEPA variants perform competitively on each dataset. Comparing pretraining datasets, ImageNet-pretrained I-JEPA outperforms its HLS variant on datasets where no multispectral information is present. Here, the missing bands are zero-padded.\end{document}